\pdfoutput=1
\documentclass[runningheads]{llncs}

\usepackage{eccv}

\usepackage{eccvabbrv}

\usepackage{times}
\usepackage{epsfig}
\usepackage{graphicx}
\usepackage{amsmath,amssymb}
\usepackage{booktabs}
\usepackage{multirow}
\usepackage{microtype}
\usepackage{algorithm}
\usepackage{algorithmic}
\usepackage[table]{xcolor}
\usepackage{wrapfig}
\usepackage{marvosym}
\usepackage{tikz}
\usetikzlibrary{arrows.meta,positioning,shapes.misc,fit,calc}
\usepackage{listings}
\usepackage{xspace}

\definecolor{customgray}{HTML}{D9D9D9}
\definecolor{customblue}{HTML}{DDE6F7}
\definecolor{cvprblue}{rgb}{0.21,0.49,0.74}

\newcommand{\ours}{\textsc{ReGRPO}}

\makeatletter
\DeclareRobustCommand\onedot{\futurelet\@let@token\@onedot}
\def\@onedot{\ifx\@let@token.\else.\null\fi\xspace}

\providecommand{\eg}{\emph{e.g}\onedot}

\makeatother

\usepackage[breaklinks,colorlinks,citecolor=eccvblue]{hyperref}

\usepackage{orcidlink}

\begin{document}
\raggedbottom

\title{ReGRPO: Reflection-Augmented Policy Optimization for Tool-Using Agents}

\titlerunning{ReGRPO: Reflection-Augmented Policy Optimization for Tool-Using Agents}

\author{Binjie Zhang \and Mike Zheng Shou\thanks{Corresponding author.}}

\authorrunning{B.~Zhang and M.~Z.~Shou}

\institute{Show Lab, National University of Singapore\\
\email{binjie97@u.nus.edu}}

\maketitle

\begin{abstract}
  Tool-augmented vision--language models (VLMs) can solve multimodal, multi-step tasks by calling external tools, yet they remain fragile in practice. Existing works have two common gaps. Supervised fine-tuning (SFT) is built mostly on successful trajectories and offers little signal for recovery after tool failures, while sparse trajectory-level RL rewards provide limited guidance on which step failed and how to repair it.
  We introduce ReGRPO (Reflection-augmented Group Relative Policy Optimization), a framework that learns reflection-guided correction in tool-using agents. ReGRPO starts with a structured reflective data engine: we execute near-miss actions to collect grounded failure observations, then build Reflection-of-Thought triplets (ErrorType, Evidence, FixPlan) paired with corrected actions for warm-start SFT. We then optimize reflection tokens and corrective actions jointly within local trajectories using group-relative advantages, and include a reflection-cost term to reduce unnecessary reflection.
  Experiments on GTA and GAIA show that, under the same backbone and tool suite, ReGRPO consistently outperforms strong open-source baselines and achieves the best results among the compared open-source controllers.
  Code and RoT data are available at \url{https://github.com/showlab/ReGRPO}.

\end{abstract}

\begin{figure*}[t]
  \begin{center}
    \includegraphics[width=\linewidth]{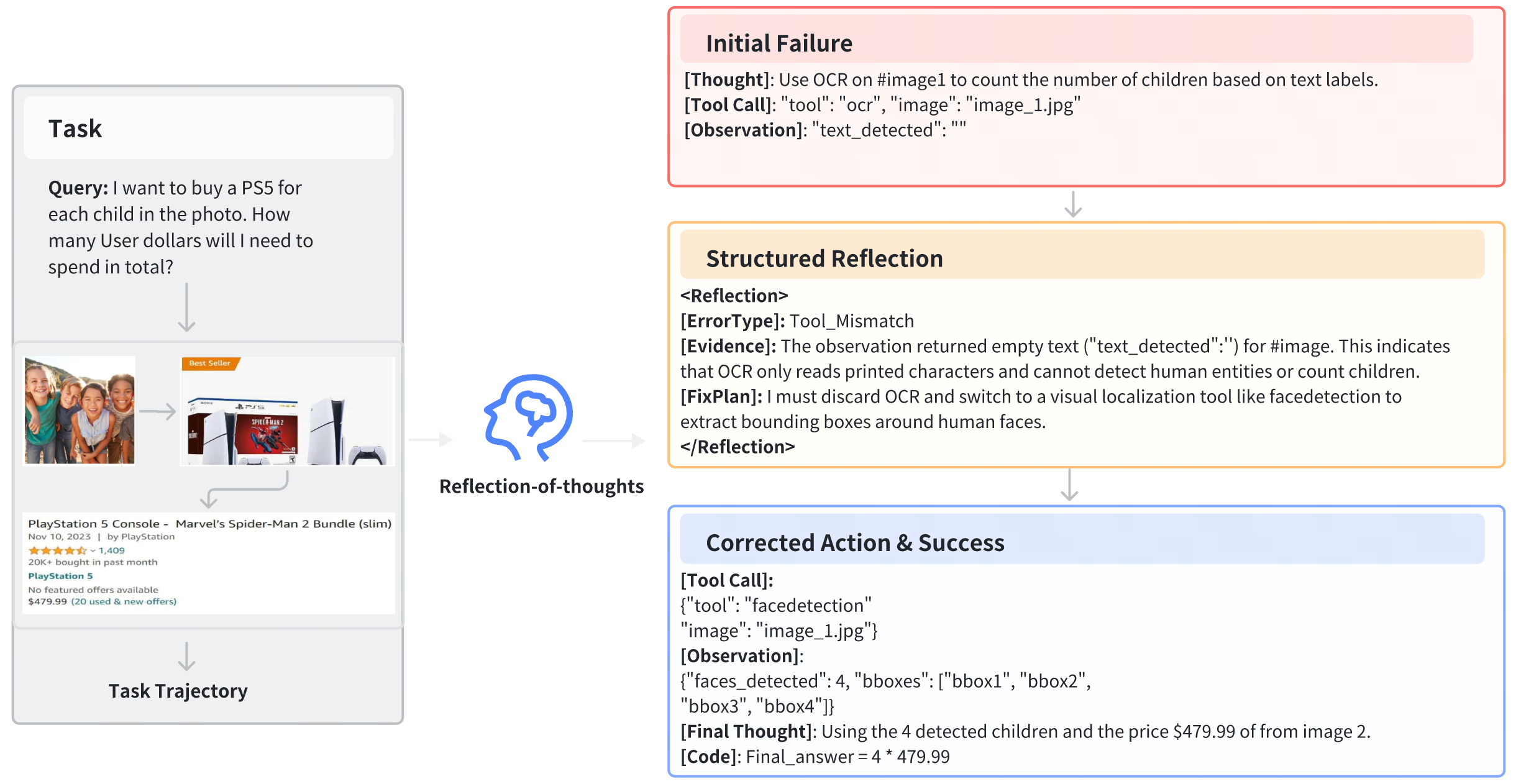}
    \caption{Pipeline of the Structured Reflective Data Engine. Given a task trajectory, we first induce a tool failure (for example, OCR on a face image returns empty text). A teacher model (GPT-4o by default) then generates a structured Reflection-of-Thought with explicit ErrorType, Evidence, and FixPlan, which explains the failure and proposes the next action (for example, switching to face detection). The agent executes the corrected action to recover and finish the task. This Error, Reflection, and Correction loop converts raw failures into grounded supervision for recovery.}
    \label{fig:mmcot_pipeline}
  \end{center}
\end{figure*}

\section{Introduction}
\label{sec:intro}

External tools such as web search, OCR, table readers, PDF parsers, code execution, and visual operators expand the capability of vision-language models (VLMs) beyond in-context prompting~\cite{suris2023vipergpt,fan2024videoagent,lu2024deepseek,guo2025deepseek}. In these settings, success depends not only on the final answer, but also on planning and executing grounded intermediate steps, including which tool to call, when to call it, and how to set its arguments from multimodal evidence.

One common approach is to train trajectory-tuned controllers on synthetic tasks and verified traces. MAT-AGENT~\cite{gao2024multi} follows this approach by generating multimodal tool-use tasks, executing tools to collect successful trajectories, and applying supervised fine-tuning (SFT) to learn tool invocation. This strategy is effective, but it also ties the controller closely to supervised traces. When the policy deviates, for example under a new PDF layout or receipt style, the training signal offers limited guidance for recovery.

A complementary direction uses self-exploration. SPORT~\cite{li2025iterative} alternates between sampling tool-use steps and verifying them, then converts rollouts into step-wise preferences without human labels. This introduces process-level feedback, but the supervision remains relatively unstructured. For example, SPORT does not explicitly encode multimodal chain-of-thought, grounding tags that link language to regions or cells, or verifier rationales. As a result, the learned preferences are often hard to interpret and can fail under distribution shift.

These approaches share two limitations. SFT-only tuning on expert trajectories tends to saturate quickly because the loss depends heavily on one teacher trace and provides weak signals for local repair. Standard reinforcement learning also provides limited recovery-oriented supervision in long-horizon tasks, because a scalar failure reward does not identify which decision should be revised.

We therefore train tool-using VLM agents with reflection-augmented reinforcement learning, where diagnostic reflections are optimized as explicit recovery signals. Instead of treating reflection as a test-time prompting heuristic, we model it as a learnable variable that links a failed action to its correction.

We introduce ReGRPO (Reflection-augmented Group Relative Policy Optimization) to realize this idea. On the data side, ReGRPO builds a Structured Reflective Data Engine that converts tool-execution failures into grounded Reflection-of-Thought (RoT) triplets. The engine perturbs MAT-AGENT trajectories~\cite{gao2024multi}, executes the perturbed actions to obtain real failure observations, and then generates structured reflections with ErrorType, Evidence, and FixPlan. On the optimization side, ReGRPO jointly optimizes reflection and correction tokens within the same local trajectory objective, which gives the policy step-level supervision beyond final success or failure. ReGRPO keeps the standard GRPO optimizer and adds structured reflection parameterization to enable end-to-end policy learning.

At inference time, ReGRPO follows the same principle that failures should trigger explicit local repair rather than blind continuation. We use a lightweight single-path, zero-verifier setup with a deterministic trigger, which opens a reflection-correction block only when needed. This design allows recovery from tool errors while keeping deployment efficient.

On GTA~\cite{wang2024gta} and GAIA~\cite{mialon2023gaia}, under the same backbone and tool suite as MAT-AGENT and SPORT, ReGRPO consistently improves overall accuracy.

In summary, this paper makes three contributions:
\begin{itemize}
  \item We build a \textit{Structured Reflective Data Engine} that converts tool-execution failures into grounded Reflection-of-Thought supervision with ErrorType, Evidence, and FixPlan, paired with corrected actions.
  \item We present \textit{ReGRPO}, which adopts the standard GRPO optimizer while adding structured reflection trajectory parameterization and a zero-verifier trigger, enabling joint optimization of reflection and correction tokens under reproducible single-path deployment.
  \item Under the same backbone and tool settings, ReGRPO achieves the \textit{strongest results} among the compared open-source controllers on GTA and GAIA.
\end{itemize}

\section{Related Work}
\label{sec:related}

\subsection{Multi-Modal Agents}

Multimodal agents~\cite{li2024llava,yin2024agent,zhang2024agentohana,wang2024mllm,liu2024llavanext} can solve complex problems via external tools and APIs. For example, CLOVA~\cite{gao2024clova} uses LLMs as controllers to compose off-the-shelf visual tools. ViperGPT~\cite{suris2023vipergpt} uses code-generation models to compose vision-and-language models into subroutines to produce results for arbitrary queries. VideoAgent~\cite{fan2024videoagent} adopts multi-step reasoning, where the agent selects tools according to intermediate observations.

Though LLM-driven methods can achieve strong results, VLM-driven agents~\cite{sasazawa2024layout,zheng2024gpt,wang2024genartist} are often more efficient for visual tasks because the controller directly consumes images or videos and tool outputs. For example, GenArtist~\cite{wang2024genartist} proposes a unified image generation and editing system coordinated by a multimodal large language model (MLLM) agent.

In addition, other works~\cite{xiong2025llava,liao2024can,wang2024divide} use models to generate AI feedback that improves performance. For example, LLaVA-Critic~\cite{xiong2025llava} presents a high-quality dataset tailored to follow instructions in complex evaluation settings, providing quantitative judgments and accompanying reasoning.

\subsection{Datasets for Tool-Using Agents}

The reasoning ability of VLM-driven agents is often weaker than that of large text-only LLMs. To bridge this gap, recent works synthesize tool-usage data to tune open-source VLMs~\cite{wang2024mllm,liu2023llava,liu2024visualagentbench}.
For example, DEDER~\cite{choiembodied} uses in-context learning to generate trajectories and distills chain-of-thought reasoning from LLMs to smaller models. Lumos~\cite{yin2024agent} converts ground-truth reasoning steps from existing benchmarks into tool-usage trajectories. TASKBENCH~\cite{shen2023taskbench} samples trajectories from pre-defined graphs.
MAT-AGENT~\cite{gao2024multi} scales up trajectory tuning for open VLM controllers with a diverse tool suite and synthetic tasks.

\subsection{Reflective and Self-Correcting Agents}

The concept of reflection---prompting a model to critique its own outputs---has been widely explored in LLMs. {Reflexion}~\cite{shinn2024reflexion} and similar frameworks~\cite{madaan2024selfrefine,chen2023teaching} use verbal feedback and episodic memory to improve performance over multiple trials at inference time. However, these methods typically treat reflection as a frozen prompting strategy or rely on external scalar feedback, without optimizing the reflection generation process itself.
In contrast, ReGRPO explicitly learns how to generate a diagnostic reflection and how that reflection guides the next corrective tool call under an explicit trigger mechanism. Rather than relying on inference-time trial-and-error alone, ReGRPO trains the model to produce grounded diagnostics that enable successful recovery, internalizing the correction loop into the policy. We evaluate this approach on GTA and GAIA to assess effectiveness and sample efficiency in tool-using settings.

\begin{figure*}[t]
  \begin{center}
    \includegraphics[width=\linewidth]{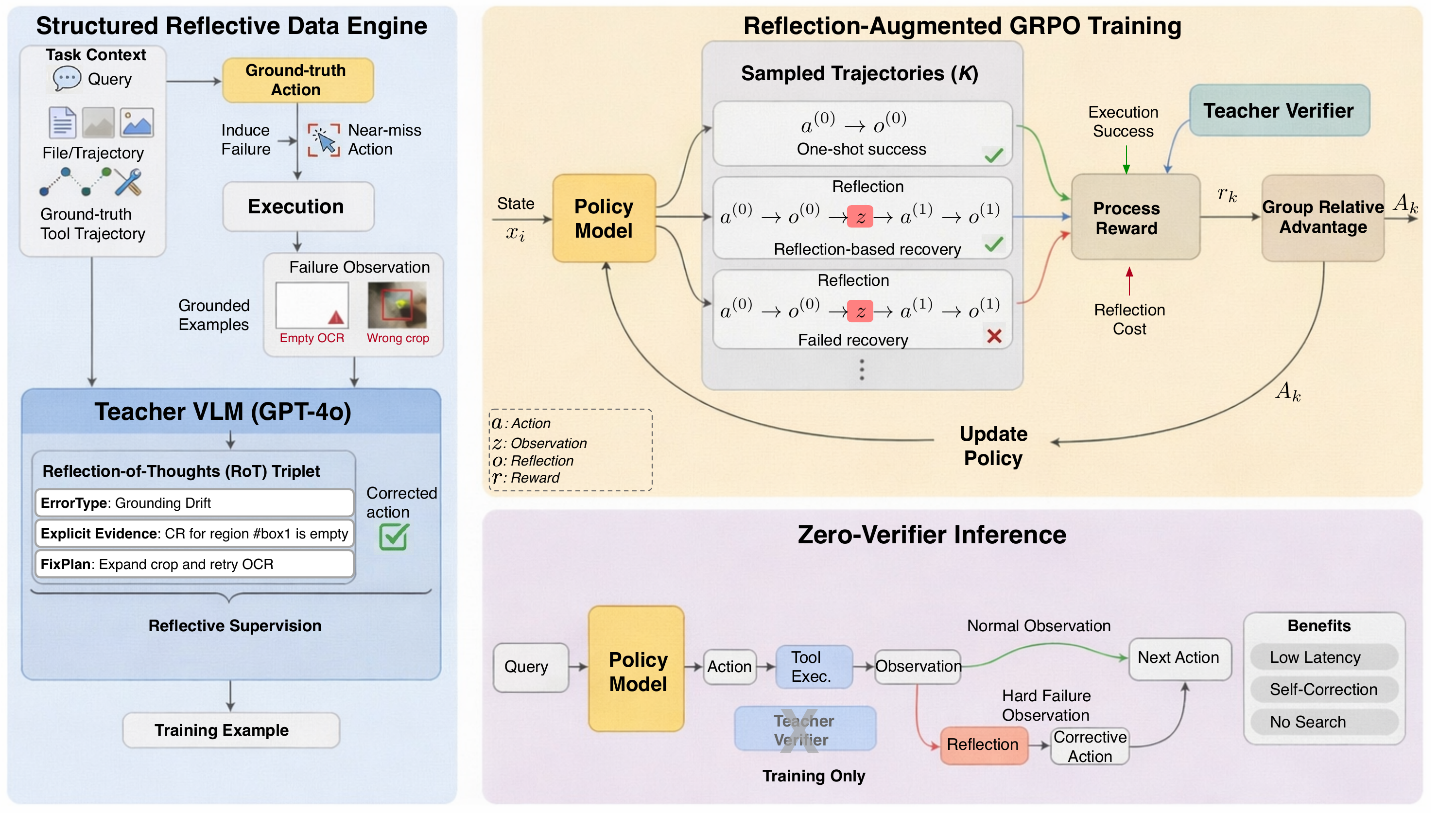}
    \caption{Overview of ReGRPO. (1) {Structured Reflective Data Engine}: from multimodal inputs and a successful action, synthesize a near-miss failure (wrong crop/tool/argument), execute it to obtain grounded failure observations (e.g., empty OCR or tool error), then use a teacher VLM (e.g., GPT-4o) to generate a structured Reflection-of-Thought triplet (ErrorType, Evidence, FixPlan). Pair the reflection with the corrected action to form reflective supervision (failure action, failure observation, reflection, corrected action), and warm-start SFT on these trajectories. (2) {ReGRPO training}: form groups of candidate local trajectories, including one-shot successes and reflection-based recoveries $a^{(0)}\!\rightarrow\!o^{(0)}\!\rightarrow\!z\!\rightarrow\!a^{(1)}\!\rightarrow\!o^{(1)}$; combine execution success, an optional teacher-derived verifier score (training only, computed deterministically from the teacher's RoT metadata with no in-loop LLM call), and reflection cost into a reward; compute group-relative advantages to update both reflection and correction tokens. (3) {Zero-Verifier Inference Stage}: single-path execution with a deterministic trigger that opens a local reflection-correction block only when failure evidence appears, enabling efficient recovery without external verifier calls.}
    \label{fig:framework}
  \end{center}
\end{figure*}

\section{Method}
\label{sec:method}

Existing training strategies for tool-using agents have two key issues. First, supervised trajectories rarely include recovery steps~\cite{gao2024multi}. As a result, the model learns only successful traces. A small mistake can then cascade without guidance on how to fix it. For example, in receipt QA, SFT teaches the correct \texttt{object\_loc} followed by OCR. At test time, a slight layout shift may lead to a nearby crop and empty OCR, but the training data provides no corrective signal (e.g., expand the box or re-localize).
Second, standard RL provides weak localization. A multi-step document QA run can fail after a wrong crop, then an empty OCR, then an incorrect answer. A final reward of 0 does not reveal which step was wrong or how to correct it. This ambiguity makes learning slow and unstable.

We introduce \textbf{ReGRPO} (Reflection-Augmented Group Relative Policy Optimization), which explicitly learns \emph{how} to generate diagnostic reflection and \emph{how} reflection guides the next corrective tool call in local trajectories.
As shown in Figure~\ref{fig:framework}, ReGRPO follows three principles.
First, it uses a {Structured Reflective Data Engine} that converts execution failures into grounded diagnostic triplets.
Second, it applies {ReGRPO} to optimize the generation of reflection steps that lead to successful corrections.
Third, it uses a Zero-Verifier Inference Stage with single-path execution and no external verifier calls.
We evaluate the resulting agent on GTA and GAIA benchmarks to assess overall tool-use performance.

The remainder of this section presents the problem setup (Sec.~\ref{subsec:problem}), the Structured Reflective Data Engine (Sec.~\ref{subsec:mmcot}), ReGRPO training (Sec.~\ref{subsec:grpo}), and the inference strategy (Sec.~\ref{subsec:inference}).

\subsection{Problem Setup}
\label{subsec:problem}

A task is given by $(Q, F)$, where $Q$ is a user query and $F$ is a set of files or images.
At step $i$, the agent observes history $h_i = \{(t_j, c_j, o_j)\}_{j=1}^{i-1}$.
Standard agents predict an action $a_i = (t_i, c_i)$, where $t_i$ is a thought and $c_i$ is a tool call.

In ReGRPO, the action space includes an additional reflection step $z_i$. When an action $a_i^{(0)}$ fails and produces observation $o_i^{(0)}$, the agent may generate reflection $z_i$ before attempting correction $a_i^{(1)}$. The local trajectory segment is
\begin{equation}
  \tau_i =
  \begin{cases} (a_i^{(0)}, o_i^{(0)}) & \text{if success} \\ (a_i^{(0)}, o_i^{(0)}, z_i, a_i^{(1)}, o_i^{(1)}) & \text{if reflection triggered}
  \end{cases}
  \label{eq:traj}
\end{equation}
The goal is to learn a policy $\pi_\theta$ that both acts and reflects to self-correct when necessary.

\subsection{Structured Reflective Data Engine}
\label{subsec:mmcot}

Existing trajectory cloning methods and outcome-based reward models provide only coarse supervision in multimodal environments. When a trajectory fails, a scalar penalty does not reveal \emph{which} step or parameter caused the error, so the model cannot learn how to repair it. To address this, we introduce a Structured Reflective Data Engine that converts failures into explicit causal evidence. Instead of treating failed actions as generic negatives---which can worsen off-policy shifts---we reformulate them as instructional ``Error-Reflection-Correction'' trajectories, giving the agent concrete recovery supervision from the outset.

\vspace{8pt}
\noindent \textbf{Failure Induction.}
To teach recovery, the model must observe concrete failure states rather than only successful traces. We therefore construct \emph{initial failures} that are realistic yet recoverable, so the agent can learn how errors arise and how to fix them.
Specifically, starting from a ground-truth step $a_i^*$ (from MM-Traj~\cite{gao2024multi}), we synthesize plausible ``near-miss'' actions $a_i^{fail}$ by perturbing tool choices or arguments (e.g., shifting a bounding box, selecting an adjacent table column, or calling a mismatched tool). We then \emph{execute} $a_i^{fail}$ in the sandbox to obtain real failure observations $o_i^{fail}$ (e.g., API exceptions, empty OCR outputs, or irrelevant crops).
These executed failures provide grounded error signals that the Reflective Data Engine links to diagnostic reflections and corrections. We organize the resulting annotations using a Reflection-of-Thought (RoT) reflection $z_i$. Each annotation is a RoT triplet $(a_i^{fail}, o_i^{fail}, z_i)$ paired with the corrected action $a_i^*$, which together explicitly encode the failure, structured reflection, and recovery target.

\vspace{8pt}
\noindent \textbf{Reflection Annotation.}
To bridge the causal gap between the faulty action $a_i^{fail}$ and the failure observation $o_i^{fail}$, we use a teacher vision-language model (GPT-4o by default) to generate a structured RoT reflection $z_i$, conditioned on the failure context $(h_i, a_i^{fail}, o_i^{fail})$. To prevent the agent from generating free-form, hallucinated excuses, $z_i$ is strictly constrained to a triplet schema:
\begin{itemize}
  \item \textbf{ErrorType}: A categorical diagnosis of the failure (e.g., \texttt{Tool\allowbreak Mismatch}, \texttt{Arg\allowbreak Invalid}, \texttt{Grounding\allowbreak Drift}, \texttt{Info\allowbreak Insufficient}).
  \item \textbf{Evidence}: A mandatory reference to the visual or textual observation that triggered the error. For instance, rather than a generic ``the tool failed'', the model must explicitly ground its reasoning: ``The OCR output for the specific region \texttt{\#box1} returned empty text, indicating an incorrect crop.''
  \item \textbf{FixPlan}: A concrete, actionable natural language strategy to correct the error and reach the target state (e.g., ``Expand the bounding box slightly to cover the text'' or ``Switch to a visual localization tool'').
\end{itemize}

\noindent \textbf{Correction Pairing and Internalization.}
Finally, the original ground-truth action $a_i^*$ is appended as the corrected action, closing the causal loop. This pipeline converts the MM-Traj dataset~\cite{gao2024multi} into a supervised corpus of augmented trajectories: $\tau_i^{reflective} = (x_i, a_i^{fail}, o_i^{fail}, z_i, a_i^*)$. During supervised fine-tuning (SFT), the model is forced to maximize the conditional likelihood $P(z_i, a_i^* \mid x_i, a_i^{fail}, o_i^{fail})$. Consequently, the agent internalizes the multimodal error diagnostic capability at the parameter level, enabling subsequent training-time exploration over reflective recoveries in ReGRPO.

\subsection{Reflection-Augmented Group Relative Policy Optimization}
\label{subsec:grpo}

\noindent \textbf{Dynamic Local Trajectory Formulation.}
For a given state context $x_i$, rather than scoring atomic actions, we form a group of $K$ candidate local trajectories $\{\tau_i^{(k)}\}_{k=1}^K$ and score each under the current policy $\pi_\theta$. The structure of each candidate trajectory depends on the intermediate environmental feedback: a candidate begins with an initial action $a_i^{(0)}$ and its observation $o_i^{(0)}$.
If the execution succeeds, the trajectory terminates early: $\tau_i = (a_i^{(0)}, o_i^{(0)})$.
Otherwise, if $o_i^{(0)}$ indicates a hard failure (e.g., execution error, empty visual crop) or low task--observation consistency, the policy generates a diagnostic reflection $z_i$ and a corrective action $a_i^{(1)}$, forming the recovery segment in \eqref{eq:traj}.

\vspace{8pt}
\noindent \textbf{Reflection-Aware Process Reward.}
To encourage meaningful exploration while penalizing infinite loops or verbosity, we define a composite reward function $R(\tau)$ that balances task success with execution efficiency:
\begin{equation}
  R(\tau) = \lambda_{\text{exec}}\,\mathbf{1}\{\text{success}\} - \eta\,C(\tau) + \lambda_{\text{val}}\,V(x_i, \tau).
  \label{eq:reward}
\end{equation}
Here, $\mathbf{1}\{\text{success}\}$ is deterministic environment feedback indicating task completion, and $C(\tau)$ penalizes unnecessary reflection length. The first two terms already form a complete \emph{verifier-free} objective. $V(\cdot)$ is an \emph{optional} training-only verifier value, computed \emph{deterministically} from the active record's RoT metadata (no model is queried inside the RL loop): it can be enabled with $\lambda_{\text{val}}>0$ for extra stabilization, or disabled with $\lambda_{\text{val}}=0$ without changing the deployment algorithm.
Default coefficients in our reported setting are $\lambda_{\text{exec}}=1.0$, $\lambda_{\text{val}}=0.3$, and $\eta=0.1$.

The term $C(\tau)$ introduces a {reflection cost penalty}. If the trajectory invokes the reflection step $z_i$, $C(\tau)$ is proportional to the token length of $z_i$; otherwise, for a one-shot success, $C(\tau) = 0$. The penalty coefficient $\eta$ forces the agent to reflect \emph{only} when strictly necessary, ensuring that the expected reward gain from a successful recovery strictly outweighs the penalty of generating additional reasoning tokens.

\textbf{Optional training-time verifier design.} When enabled, the verifier value $V(x_i,\tau)$ is computed \emph{deterministically} from each candidate and the active record's RoT metadata; \textbf{no GPT-4o (or any LLM) is queried inside the RL loop}. We derive three subscores in $[0,1]$ by signature matching and a grounded-reflection check rather than by model scoring:
\begin{itemize}
  \item {Plan validity} $s_p$: $s_p=1$ iff the candidate's normalized primary tool and first argument match the stored corrected-action signature, else $0$.
  \item {Answer consistency} $s_a$: $s_a=1$ iff the candidate's replay succeeds (its terminal action agrees with the group's correct answer), else $0$.
  \item {Grounding} $s_g$: $s_g=1$ iff the candidate carries a reflection whose evidence is text-grounded in the stored failure observation \emph{and} $s_p>0$, else $0$. There is intentionally no fallback to $s_p$ for reflection-less candidates, so $V$ rewards grounded reflection above a bare plan repair.
\end{itemize}
The overall score is the weighted sum $V = w_a s_a + w_g s_g + w_p s_p$, with each subscore clamped to $[0,1]$ and weights set to emphasize grounding ($w_g \ge w_a, w_p$); we use $(w_a,w_g,w_p)=(0.25,0.50,0.25)$ by default. Because $V$ is a function of the candidate and the metadata, it is used only as a training-time reward-shaping signal and never as a deployment-time call. A GPT-4o teacher is used \emph{only offline}, to synthesize the RoT reflections in the data engine (Sec.~\ref{subsec:mmcot}); it is never queried during RL or at inference. Full coefficient ranges, the deterministic subscore definitions, and the selection protocol are provided in Appendix~\ref{sec:verifier_prompt}.

\subsection{Training Objective}

\noindent \textbf{Structured Reflective Data in Supervised Fine-Tuning (warm start).}
We first teach the model to diagnose and correct failures with explicit supervision, so Reinforcement Learning (RL) can focus on \emph{how} reflection should be written under explicit trigger gating rather than learning recovery from scratch.
Given the failure context $(x_i, a_i^{fail}, o_i^{fail})$, we maximize the likelihood of the structured reflection and corrected action:
\begin{equation}
  \mathcal{L}_{\mathrm{SFT}} = -\mathbb{E}_{(x_i,a_i^{fail},o_i^{fail},z_i,a_i^*)} \left[ \log P_\theta(z_i, a_i^* \mid x_i, a_i^{fail}, o_i^{fail}) \right].
\end{equation}
This stage teaches the model \emph{what} to diagnose and \emph{how} to fix errors, providing a strong initialization for subsequent RL.

\begin{figure}[!t]
  \centering
  \includegraphics[width=1\textwidth]{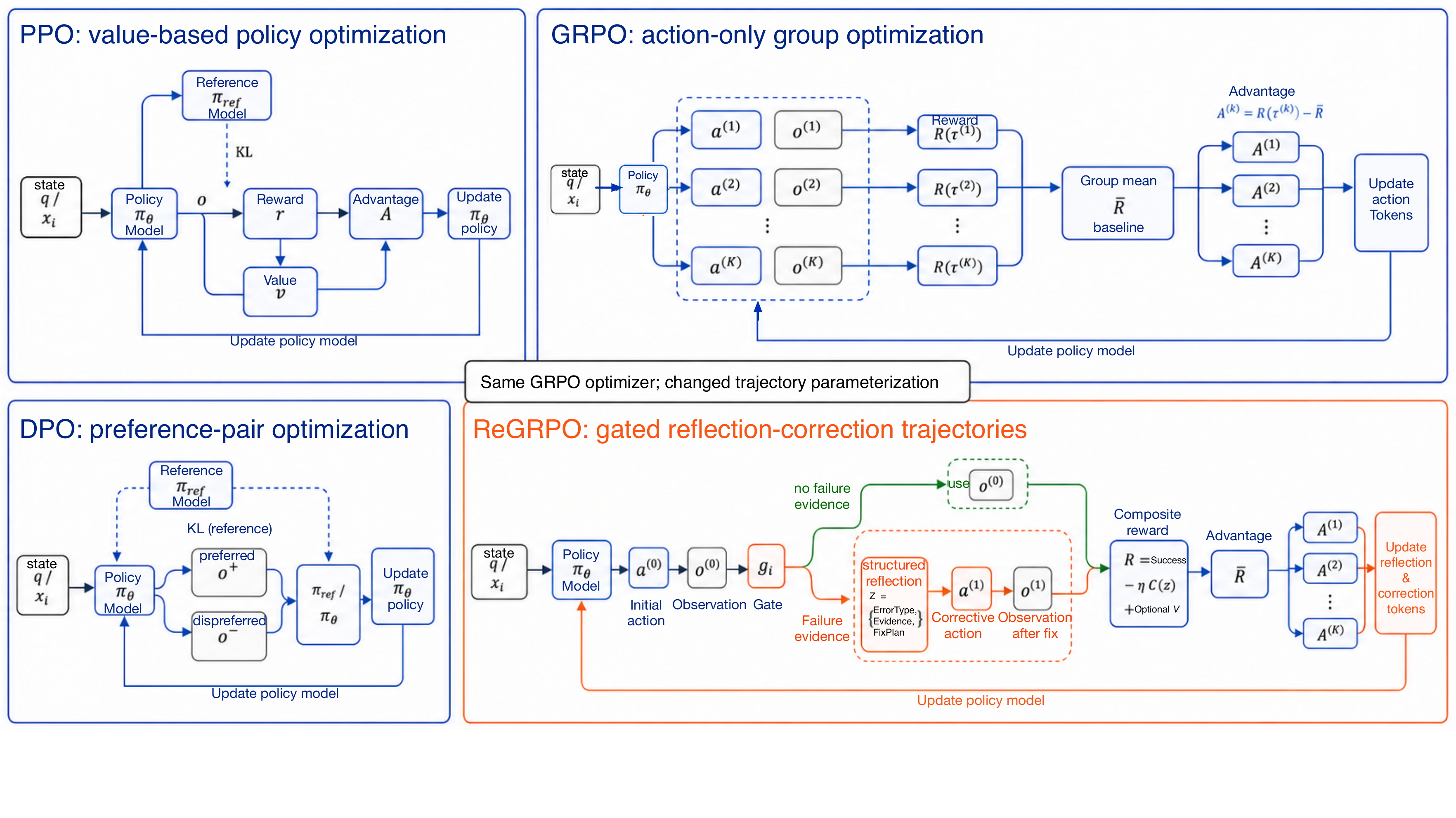}
  \caption{Comparison of PPO~\cite{schulman2017proximal}, DPO~\cite{rafailov2023direct}, GRPO~\cite{shao2024deepseekmath}, and ReGRPO (ours). PPO and DPO optimize actions or preferences without treating reflection as a decision variable; GRPO reduces variance via group-relative rewards; ReGRPO further includes reflection in the optimized trajectory to provide stronger recovery-oriented supervision for failed steps.}
  \label{fig:comparision_regrpo}

\end{figure}

\vspace{8pt}
\noindent \textbf{Structured Reflective Data in Policy Optimization.}
SFT on RoT data teaches the model to fix common tool-call errors (e.g., crop shift, wrong region, mismatched tool arguments), but real-world failures are more diverse. Therefore, to generalize beyond curated patterns, the agent must self-explore to discover new failure modes and recovery strategies. This motivates an RL stage to expand coverage and improve robustness.

However, PPO~\cite{schulman2017proximal} and DPO~\cite{rafailov2023direct} optimize actions or preference pairs without modeling reflection as a decision variable. In long-horizon tool use, a scalar success/failure signal does not indicate \emph{which} step failed or what correction would fix the trajectory. Moreover, DPO relies on preference data that is difficult to collect for multi-step tool executions, and PPO can be sample-inefficient under sparse rewards.
In contrast, GRPO~\cite{shao2024deepseekmath} optimizes relative rewards within a group of sampled trajectories, producing lower-variance advantages that are well suited to sparse, delayed rewards. By comparing recoveries against nearby failures, GRPO yields more informative gradients than absolute-reward optimization.

As illustrated in Figure~\ref{fig:comparision_regrpo}, ReGRPO extends GRPO by making reflection $z_i$ part of the optimized trajectory. This ties the advantage signal directly to diagnostic tokens, so the policy learns \emph{which} reflections are useful and \emph{how much} reflection is worth generating. Consequently, reflection quality and brevity become explicitly optimized rather than a fixed prompting heuristic.

\vspace{8pt}
\noindent \textbf{Reflection-Driven Advantage and Optimization.}
Within the sampled group, we compute the baseline as the mean reward $\bar{R}_i = \frac{1}{K}\sum_{k=1}^K R(\tau_i^{(k)})$, and extract the advantage for each trajectory: $A_i^{(k)} = R(\tau_i^{(k)}) - \bar{R}_i$.

For two trajectories from the same state context, an unrecovered failure $\tau^{-}$ and a recovered trajectory $\tau^{+}$, the reward gap is $\Delta R = R(\tau^{+}) - R(\tau^{-})$.
When recovery succeeds and reflection remains concise, $\Delta R$ is typically positive, so $A(\tau^{+})>A(\tau^{-})$ within the same sampled group. Importantly, this remains true in the verifier-free setting ($\lambda_{\text{val}}=0$), where improvements come entirely from reflection-conditioned recovery and reflection-cost control.

The ReGRPO objective is optimized as:
$$ \mathcal{L}_{\mathrm{ReGRPO}} = -\mathbb{E}_{x_i, \tau \sim \pi_\theta} \left[ \frac{1}{K} \sum_{k=1}^K A_i^{(k)} \log \pi_\theta(\tau_i^{(k)}\mid x_i) \right] + \beta\,\mathbb{D}_{\mathrm{KL}}\!\big(\pi_{\theta}\|\pi_{\text{ref}}\big) $$
Since the generation probability of structured reflection $\log \pi_\theta(z_i \mid x_i, a_i^{(0)}, o_i^{(0)})$ is factorized within the trajectory likelihood $\log \pi_\theta(\tau_i^{(k)}\mid x_i)$, relative advantages directly scale gradients on reflection tokens. ReGRPO therefore uses the standard GRPO optimizer and advantage estimator; the contribution is the structured reflection trajectory design and training/deployment protocol, not a new value estimator. Detailed quantitative results are presented in Sec.~\ref{sec:exp}, and extended diagnostics are reported in the appendix.

\subsection{Zero-Verifier Inference Stage}
\label{subsec:inference}

At deployment, we use single-path inference without external verifier calls. A deterministic gate decides whether the policy should open one local reflection-correction block.
Tool outputs are first normalized into a canonical schema
\begin{equation}
  \hat{o}_i = \{\texttt{status},\texttt{payload},\texttt{meta}\},
\end{equation}
where \texttt{status} stores backend error codes and \texttt{payload} stores normalized content.
We then use a minimal trigger
\begin{equation}
  g_i = \mathbf{1}\{\texttt{ToolError}(\hat{o}_i) \lor \texttt{EmptyObs}(\hat{o}_i) \lor u_i < \kappa_i\},
\end{equation}
with policy confidence
\begin{equation}
  u_i = \exp\!\left(\frac{1}{|a_i^{(0)}|}\sum_j \log \pi_\theta(a_{i,j}^{(0)}\mid x_i,h_i)\right).
\end{equation}
The first two terms capture explicit runtime failures. The confidence term is a lightweight uncertainty proxy: it measures token-level confidence for the predicted action sequence. We therefore use it only as a trigger heuristic for potential silent failures.
To avoid per-tool offline calibration, we use an online adaptive threshold computed from the current trajectory:
\begin{equation}
  \kappa_i = \frac{1}{\max(1,i-1)}\sum_{j=1}^{i-1} u_j.
\end{equation}
For $i=1$, we disable the confidence trigger and rely on hard-failure checks only.
If $g_i=0$, inference continues with the next standard action. If $g_i=1$, the policy executes one local block $a_i^{(0)}\rightarrow o_i^{(0)}\rightarrow z_i\rightarrow a_i^{(1)}\rightarrow o_i^{(1)}$, with at most one reflection-correction block per step.
This design keeps inference deterministic and lightweight, while reducing manual feature engineering and per-tool tuning.

Pseudo-code for this trigger is provided in the Appendix.

\section{Experiments}
\label{sec:exp}

\subsection{Benchmarks and Metrics}
We evaluate ReGRPO on two multimodal tool-use benchmarks in the MAT-AGENT setting~\cite{gao2024multi}, where agents must reason over images and documents with real tools.

\noindent\textbf{GTA Dataset.}
GTA~\cite{wang2024gta} contains 229 tasks paired with 252 images.
Each task requires 2--8 tool-use steps (typically 2--4) and tests visual perception, local operations (e.g., reading receipts or charts), and short reasoning chains over screenshots and UI-like images.

\noindent\textbf{GAIA Dataset.}
GAIA~\cite{mialon2023gaia} is a document-centric benchmark with 446 tasks over 109 files in PPTX, PDF, and XLSX formats.
Tasks are grouped into three difficulty levels and often require multiple tool calls for document understanding, web navigation, logical reasoning, and summarization.

\begin{table*}[t]
  \centering
  \caption{Main comparison on GTA and GAIA under a unified single-path, zero-verifier inference protocol.
  The default verifier-free ReGRPO ($\lambda_{\text{val}}=0$) achieves the best performance among the evaluated open-source controllers on both benchmarks.}
  \label{tab:main_res}
  \scriptsize
  \resizebox{1\textwidth}{!}{%
    \begin{tabular}{l|c|ccc|cccc}
      \hline
      \multirow{2}{*}{\textbf{Method}} & \multirow{2}{*}{\textbf{Controller}} & \multicolumn{3}{c|}{\textbf{GTA}} & \multicolumn{4}{c}{\textbf{GAIA}} \\
      &  & \textit{ToolAcc} & \textit{CodeExec} & \textit{AnsAcc} & \textit{Level 1} & \textit{Level 2} & \textit{Level 3} & \textit{AnsAcc}   \\
      \hline
      \rowcolor{customgray} \multicolumn{9}{c}{\textit{Closed-source Controller}} \\ \hline
      Lego Agent       & GPT-4           & -      & -       & 46.59 & -      & -      & -      & -          \\
      Lego Agent       & GPT-4o          & -      & -       & 41.52 & -      & -      & -      & -         \\
      Warm-up Agent    & GPT-4-turbo     & -      & -       & -     & 30.20  & 15.10  & 0.00   & 17.60    \\
      HF Agent         & GPT-4o          & 63.41  & 95.12   & 57.05 & 47.17  & 31.40  & 11.54  & 33.40    \\
      HF Agent         & GPT-4o-mini     & 56.10  & 100.00  & 57.69 & 33.96  & 27.91  & 3.84   & 26.06    \\ \hline
      \rowcolor{customgray} \multicolumn{9}{c}{\textit{Open-Source Controller}} \\ \hline
      HF Agent         & LLaVA-NeXT-8B   & 14.97  & 25.08   & 14.10 & 9.43   & 1.16   & 0.00   & 3.64      \\
      HF Agent         & InternVL2-8B    & 36.75  & 52.18   & 32.05 & 7.55   & 4.65   & 0.00   & 4.85     \\
      HF Agent         & MiniCPM-V-8.5B  & 36.59  & 56.10   & 33.97 & 13.21  & 5.81   & 0.00   & 7.27     \\
      HF Agent         & Qwen2-VL-7B     & 44.85  & 65.19   & 42.31 & 16.98  & 8.14   & 0.00   & 9.70     \\
      T3-Agent         & MAT-MiniCPM-V-8.5B & 65.85  & 80.49   & 52.56 & 26.42  & 11.63  & {3.84}   & 15.15   \\
      T3-Agent         & MAT-Qwen2-VL-7B & 64.63  & 84.32   & 53.85 & 26.42  & 15.12  & {3.84}   & 16.97    \\
      SPORT Agent         & Tuned-Qwen2-VL-7B & {72.41}  & {91.87}   & {60.26} & {35.85}  & {16.28}  & {3.84}   & {20.61}    \\ \hline
      \rowcolor{customblue} \multicolumn{9}{c}{\textit{Ours}} \\ \hline
      \textbf{ReGRPO (default, $\lambda_{\text{val}}=0$)}     & MAT-Qwen2-VL-7B & \textbf{76.35} & \textbf{93.77} & \textbf{67.66} & \textbf{39.02} & \textbf{18.71} & \textbf{4.89} & \textbf{23.35}   \\
      \hline
    \end{tabular}%
  }
  \vspace{-16pt}
\end{table*}

\noindent\textbf{Metrics.}
We report standard accuracy metrics: \emph{AnsAcc} (answer accuracy), \emph{ToolAcc} (tool-call validity), and \emph{CodeExec} (execution success rate).
Unless noted otherwise, all main-text comparisons use zero-verifier deployment, with no external verifier calls at test time.

\subsection{Baselines}

We compare ReGRPO with both closed-source and open-source controllers.

\noindent\textbf{Closed-source agents.}
We report GPT-4/4o-based results from prior work, including Lego Agent and Warm-up Agent~\cite{mialon2023gaia,wang2024gta}, and HF Agents powered by GPT-4o and GPT-4o-mini.
These models provide strong proprietary references.

\noindent\textbf{Open-source agents.}
We include HF Agents based on LLaVA-NeXT-8B, InternVL2-8B, MiniCPM-V-8.5B, and Qwen2-VL-7B.
We also evaluate MAT-AGENT (T3-Agent) with MAT-MiniCPM-V-8.5B and MAT-Qwen2-VL-7B, and SPORT Agent with a tuned Qwen2-VL-7B controller.
For fair internal comparisons, all ablations use the same Qwen2-VL-7B controller and tool suite.

\noindent\textbf{Our model.}
We use the same Qwen2-VL-7B backbone and toolset as SPORT~\cite{li2025iterative} to isolate the effect of our method.
We compare controlled variants under matched training settings (same total updates, rollout budget, and token budget):
(1) {MAT-AGENT}: Fine-tuned on MM-Traj without RL;
(2) {+RoT SFT}: SFT with structured reflective data only;
(3) {+Optional Verifier Distill}: adds verifier-aware distillation but no RL;
(4) {GRPO-only}: GRPO on actions only (no reflection tokens);
(5) {GRPO + Free-form Reflection}: allows unstructured reflection text without schema constraints;
(6) {ReGRPO core (default, verifier-free RL)}: structured reflection with $\lambda_{\text{val}}=0$ during RL;
(7) {ReGRPO + optional verifier reward}: additive shaping with the deterministic, metadata-derived verifier value ($\lambda_{\text{val}}>0$).
All variants use {Single-Path, Zero-Verifier Inference} at test time.

\subsection{Implementation Details}
\label{sec:impl}

\noindent \textbf{Model and optimization.}
We use Qwen2-VL-7B as the controller.
The vision encoder and visual token compressor are frozen.
We fine-tune the language model with LoRA~\cite{hu2021lora} (rank 32), applied to query, key, and value projections in all self-attention layers.
We optimize with AdamW and cosine learning-rate decay, using a base learning rate of $1.0\times10^{-6}$ and batch size 2 per device.

\noindent \textbf{Training stages.}
Our default pipeline has two mandatory stages:
(1) \emph{SFT warm start} on our Structured Reflective Data (triplets);
(2) \emph{ReGRPO process RL} on a mixture of offline groups and online self-exploration groups.
Optionally, we insert verifier-aware distillation between (1) and (2), where the controller predicts the deterministic, metadata-derived verifier subscores for better calibration and more stable initialization.
ReGRPO uses the reflection-aware reward in Eq.~\ref{eq:reward}, with a reflection-cost penalty $\eta=0.1$.
By default, we set $\lambda_{\text{val}}=0$ (verifier-free RL); runs with $\lambda_{\text{val}}>0$ using the deterministic, metadata-derived verifier value are reported as additive variants.
For controlled comparisons, variants with and without distillation use identical optimizer settings and matched update steps to avoid gains from extra training budget.

\begin{table*}[t]
  \centering
  \caption{Ablation of ReGRPO under the same backbone and tool settings. The verifier-free core pipeline (RoT + ReGRPO, $\lambda_{ \text{val}}=0$) reaches $67.66/23.35$ GTA/GAIA AnsAcc, and adding the optional teacher-derived verifier reward further improves to $68.49/24.01$. Most gains come from the RoT SFT + structured RL combination, while verifier signals act as optional additive improvements.}
  \label{tab:ablation}
  \scriptsize
  \resizebox{1\textwidth}{!}{%
    \begin{tabular}{l|ccc|ccc|cccc}
      \hline
      \multirow{2}{*}{\textbf{Method}} &
      \multicolumn{3}{c|}{\textbf{Module}} &
      \multicolumn{3}{c|}{\textbf{GTA}} &
      \multicolumn{4}{c}{\textbf{GAIA}} \\
      & \textbf{SFT Data} & \textbf{RL Alg.} & \textbf{Reflect.}
      & \textit{ToolAcc} & \textit{CodeExec} & \textit{AnsAcc}
      & \textit{Level 1} & \textit{Level 2} & \textit{Level 3} & \textit{AnsAcc} \\
      \hline

      MAT-AGENT~\cite{gao2024multi} & MM-Traj.  & - & - & 64.63  & 84.32   & 53.85 & 26.42  & 15.12  & {3.84}   & 16.97 \\
      +RoT (SFT) & RoT & - & \checkmark & 68.73 & 87.92 & 58.59 & 30.58 & 15.61 & 4.17 & 19.03 \\
      +Optional Verifier Distill (no RL) & RoT & - & \checkmark & 69.84 & 88.41 & 59.72 & 31.04 & 15.94 & 4.17 & 19.84 \\
      GRPO-only (w/o RoT) & MM-Traj. & GRPO & - & 71.34 & 90.23 & 64.51 & 36.79 & 18.12 & 3.84 & 18.92 \\
      GRPO + Free-form Reflection & RoT & GRPO & \checkmark & 73.05 & 91.02 & 65.34 & 37.26 & 18.02 & 4.17 & 21.38 \\
      \textbf{ReGRPO core (default, $\lambda_{\text{val}}=0$)} & RoT & ReGRPO & \checkmark & 76.35 & 93.77 & 67.66 & 39.02 & 18.71 & 4.89 & 23.35 \\
      + optional verifier reward (deterministic) & RoT & ReGRPO & \checkmark & 77.26 & 94.91 & 68.49 & 40.09 & 19.67 & 5.32 & 24.01 \\
      \hline
    \end{tabular}%
  }
  \vspace{-16pt}
\end{table*}

\begin{figure*}[p]
  \centering
  \scriptsize
  \setlength{\emergencystretch}{3em}
  \setlength{\tabcolsep}{3pt}
  \definecolor{codebg}{HTML}{F4F4F6}
  \newcommand{\cardwidth}{\dimexpr\linewidth-2\fboxsep-6pt\relax}
  \newcommand{\panelbar}[1]{\noindent\colorbox{customblue}{\parbox{\cardwidth}{\footnotesize\textbf{#1}}}\par\smallskip}
  \newcommand{\subbar}[2]{\par\smallskip\noindent\colorbox{#1}{\parbox{\cardwidth}{\scriptsize\textbf{#2}}}\par\smallskip}
  \newcommand{\codecard}[2]{\noindent\colorbox{codebg}{\parbox{\cardwidth}{\scriptsize\textbf{#1}\par\smallskip{\ttfamily\linespread{1.18}\selectfont #2\par}}}\par\smallskip}
  \begin{tikzpicture}
    \node[draw, rounded corners=6pt, inner sep=6pt] {
      \begin{minipage}{0.95\linewidth}
        \begin{tabular}{p{0.98\linewidth}}
          \panelbar{(a) MAT SFT vs.\ RoT training example~\textnormal{\scriptsize(verbatim record \texttt{0FLZe2lb\_rot\_s0\_GroundingDrift})}}
          {\scriptsize \textbf{MAT SFT} has top-level fields \texttt{id}, \texttt{image}, \texttt{answer}, \texttt{conversations}. \textbf{RoT} keeps \texttt{id}/\texttt{image}/\texttt{conversations} and adds a \texttt{reflection} triplet (\texttt{error\_type}, \texttt{evidence}, \texttt{fix\_plan}) plus a \texttt{corrected\_action}.\par}
          \smallskip
          \codecard{MAT instance (abbreviated)}{%
            \{"id":"0FLZe2lb",\ "image":"coco/.../000000043093.jpg",\\
            \ \ "answer":"...smoothie calorie estimate...",\\
            \ \ "conversations":[\{"role":"user",...\},\{"role":"assistant",...\}]\}}
          \codecard{RoT instance (abbreviated)~\textnormal{\textcolor{cvprblue}{\scriptsize(blue = fields RoT adds over MAT)}}}{%
            \{"id":"0FLZe2lb\_rot\_s0\_GroundingDrift",\\
            \ \ "image":"coco/.../000000043093.jpg",\\
            \textcolor{cvprblue}{\ \ "failed\_action":"visualizer(q='fruits/utensils on the table')",}\\
            \textcolor{cvprblue}{\ \ "reflection":\{"error\_type":"GroundingDrift",}\\
            \textcolor{cvprblue}{\ \ \ \ \ \ "evidence":"cutting board, banana, knife",}\\
            \textcolor{cvprblue}{\ \ \ \ \ \ "fix\_plan":"re-ground to the smoothie glass"\},}\\
            \textcolor{cvprblue}{\ \ "corrected\_action":"visualizer(q='ingredients in this smoothie')"\}}}
          \\[0.9em]

          \panelbar{(b) Mechanism-aligned reflection~\textnormal{\scriptsize(real \texttt{(ErrorType, Evidence, FixPlan)} triplet)}}
          {\scriptsize
            \textbf{Task.} ``How many calories are in the smoothie shown in this image?'' The image shows a smoothie glass beside a cutting board with a sliced banana, a knife, and granola.\par\smallskip
            \textbf{Silent failure (image grounding).} The perturbed action asks \texttt{visualizer} about the \emph{fruits and utensils on the table}; the tool returns a valid, non-empty description (``a wooden cutting board with a sliced banana, a knife, \ldots; the smoothie glass is not described''), so no explicit tool error is raised. Low policy confidence (\(u_i<\kappa_i\)) opens one local repair block.\par\smallskip
            \textbf{Reflection (ReGRPO schema, verbatim).} \texttt{ErrorType=GroundingDrift}; \texttt{Evidence="wooden cutting board with a sliced banana, a knife"}; \texttt{FixPlan="refocus from the table/cutting board to the smoothie glass and ask about its contents"}. The corrected action re-grounds the query to the smoothie itself.\par\smallskip
            \textbf{Takeaway.} This real record exposes the full Error$\rightarrow$Reflection$\rightarrow$Correction chain that ReGRPO optimizes, under the same single-path zero-verifier deployment.\par}
          \\[0.9em]

          \panelbar{(c) Inference Path Comparison}
          \codecard{Sample content}{%
            tools: visualizer\\
            files: coco/train2017/000000043093.jpg\\
            dialogs[0]: \{"role":"user","content":"How many calories are in the smoothie shown in this image?"\}}
          {\scriptsize\textit{Note: abbreviated from the verbatim RoT record; tool outputs are shortened for readability.}\par}
          \subbar{customgray}{Baseline path (no local repair block)}
          {\scriptsize
          \begin{enumerate}\setlength{\itemsep}{1pt}\setlength{\topsep}{1pt}\setlength{\parskip}{0pt}
            \item \textbf{Step 1}: \texttt{visualizer(image, q="fruits/utensils on the table")} $\rightarrow$ ``cutting board, sliced banana, knife, granola; smoothie glass not described''.
            \item \textbf{Step 2}: estimate calories from the off-target items on the board $\rightarrow$ ungrounded ingredient list (wrong basis).
          \end{enumerate}}
          \subbar{customblue}{ReGRPO path (single-path, zero-verifier)}
          {\scriptsize\sloppy
          \begin{enumerate}\setlength{\itemsep}{1pt}\setlength{\topsep}{1pt}\setlength{\parskip}{0pt}
            \item \textbf{Step 1}: same first call; the description omits the target smoothie ($g_i=0$ so far).
            \item \textbf{Step 1 (gate)}: the target object is missing from the observation, so confidence is low (\(u_i<\kappa_i\)); set $g_i=1$.
            \item \textbf{Local block}: $a_i^{(0)}\!\rightarrow\!o_i^{(0)}\!\rightarrow\!z_i\!\rightarrow\!a_i^{(1)}\!\rightarrow\!o_i^{(1)}$. Reflection emits \texttt{ErrorType=GroundingDrift}, \texttt{Evidence=cutting board / banana / knife}, and \texttt{FixPlan=re-ground to the smoothie glass}.
            \item \textbf{Step 2}: corrected action \texttt{visualizer(image, q="ingredients in this smoothie")} $\rightarrow$ grounded smoothie contents, then a grounded calorie estimate.
          \end{enumerate}}
          \\
        \end{tabular}
      \end{minipage}
    };
  \end{tikzpicture}
  \caption{Figure-level evidence for ReGRPO, instantiated on a verbatim synthesized RoT record (\texttt{0FLZe2lb\_rot\_s0\_GroundingDrift}). (a) RoT augments the MAT SFT format with explicit reflective fields (a \texttt{reflection} triplet and a \texttt{corrected\_action}). (b) The real \texttt{(ErrorType, Evidence, FixPlan)} reflection diagnoses a silent grounding failure---the tool answers about the cutting board rather than the smoothie---and prescribes a re-grounding fix. (c) Inference-path comparison contrasts a brittle baseline route that estimates calories from off-target items with a ReGRPO route where one confidence trigger opens a single local reflection--correction block that re-grounds to the smoothie.}
  \label{fig:traj_compare}
\end{figure*}

\subsection{Quantitative Results}
\label{sec:main-results}

Table~\ref{tab:main_res} summarizes results on GTA and GAIA~\cite{wang2024gta,mialon2023gaia}.
Under the same single-path, zero-verifier setup, default ReGRPO ($\lambda_{\text{val}}=0$) gives the strongest results among the compared open-source controllers.
On GTA, it reaches $76.35$ ToolAcc and $67.66$ AnsAcc, improving over SPORT~\cite{li2025iterative} by $+3.94$ ToolAcc and $+7.40$ AnsAcc. The larger AnsAcc gain suggests that reflection helps end-to-end reasoning beyond tool-call validity alone.
On GAIA, default ReGRPO raises overall AnsAcc to $23.35$ ($+2.74$ over SPORT), indicating stronger results on document-centric multi-step tasks even without verifier reward.
Adding the optional teacher-derived verifier reward further improves performance to $68.49/24.01$ GTA/GAIA AnsAcc in Table~\ref{tab:ablation}, while the main gains are already achieved in verifier-free training and deployment.
Across both datasets, the results support our hypothesis that structured reflections improve tool-grounded reasoning. Multi-seed statistics and extended analyses are provided in the appendix.

\subsection{Ablation Studies}

Table~\ref{tab:ablation} reports a controlled ablation in which we progressively add reflective supervision and reflection-aware optimization.

\vspace{8pt}
\noindent \textbf{Contribution coverage.}
Our experimental design covers these method components: (1) the \emph{Structured Reflective Data Engine} is evaluated by MAT-AGENT $\rightarrow$ +RoT SFT and the data-format comparison in Figure~\ref{fig:traj_compare}; (2) the \emph{GRPO-based reflection protocol} is demonstrated by comparing GRPO-only and free-form reflection against structured ReGRPO under the same settings; (3) \emph{deterministic zero-verifier trigger gating} is assessed under single-path, zero-verifier deployment, with extended trigger diagnostics in the appendix.

\vspace{8pt}
\noindent \textbf{Baseline and reflective SFT gains.}
Starting from MAT-AGENT (MM-Traj, no RL), performance is $53.85$ GTA AnsAcc and $16.97$ GAIA AnsAcc. Replacing SFT data with RoT (+RoT SFT) raises GTA AnsAcc to $58.59$ and GAIA AnsAcc to $19.03$ ($+4.74$ GTA / $+2.06$ GAIA), showing that explicit Error-Reflection-Correction supervision improves trajectory quality before RL.

\vspace{8pt}
\noindent \textbf{Optional verifier distillation contributes modestly.}
Adding optional verifier distillation on top of RoT SFT further improves performance to $59.72$ GTA AnsAcc and $19.84$ GAIA AnsAcc (an additional $+1.13$ GTA / $+0.81$ GAIA over +RoT SFT). This suggests the deterministic verifier subscores help intermediate decisions, but alone they do not close the gap to RL-based methods.

\vspace{8pt}
\noindent \textbf{Action-only RL vs. reflection-aware variants.}
GRPO-only (without RoT reflection tokens in policy optimization) reaches $64.51$ GTA AnsAcc and $18.92$ GAIA AnsAcc, indicating that RL exploration is effective on GTA but more limited on GAIA. Allowing free-form reflection improves results to $65.34$ GTA AnsAcc and $21.38$ GAIA AnsAcc ($+0.83$ GTA / $+2.46$ GAIA over GRPO-only), suggesting reflection text helps, while unstructured reflection remains less efficient and less consistent.

\vspace{8pt}
\noindent \textbf{Effect of structured ReGRPO and optional verifier reward.}
Switching from free-form reflection to structured ReGRPO with $\lambda_{\text{val}}=0$ yields $67.66$ GTA AnsAcc and $23.35$ GAIA AnsAcc ($+2.32$ GTA / $+1.97$ GAIA), showing that schema-constrained reflection and reflection-correction coupling add gains even without verifier reward in RL. Adding the optional teacher-derived verifier reward further improves to $68.49$ GTA AnsAcc and $24.01$ GAIA AnsAcc ($+0.83$ GTA / $+0.66$ GAIA over ReGRPO core), indicating that verifier reward is useful but not essential.

\vspace{8pt}
\noindent \textbf{Key insight.}
The ablation shows complementary effects: (1) RoT data provides a strong starting point, (2) structured reflection-aware RL adds consistent gains on top of that initialization, and (3) verifier signals provide modest additive shaping rather than core capability. ReGRPO therefore remains effective in a verifier-free default setting during both training ($\lambda_{\text{val}}=0$) and inference (zero-verifier single-path execution). We further verify that ReGRPO preserves the base model's VQA ability (Appendix Sec.~\ref{sec:base_vqa}).

\subsection{Qualitative Analysis}
Figure~\ref{fig:traj_compare} presents three views of the mechanism, instantiated on a verbatim synthesized RoT record (a \texttt{GroundingDrift} case). Panel (a) shows that RoT extends the MAT SFT format with explicit reflective fields (a \texttt{reflection} triplet and a \texttt{corrected\_action}), so supervision covers failure diagnosis and correction rather than tool calls alone. Panel (b) gives case-level evidence aligned with Sec.~\ref{subsec:inference}: the tool returns a valid but off-target description (it answers about the cutting board, not the smoothie), so no explicit error is raised, and low confidence (\(u_i<\kappa_i\)) triggers a structured reflection whose schema (ErrorType, Evidence, FixPlan) re-grounds the query rather than adding generic free-form text. Panel (c) shows the same mechanism as paths: the baseline route estimates from off-target items, while ReGRPO opens exactly one local block at the uncertain step and rewrites the action to re-ground on the target object. The full record is given in Appendix Table~\ref{tab:rot_b4}.

\section{Conclusion}
\label{sec:conclusion}

We presented \textbf{ReGRPO}, a reflection-augmented framework for training tool-using vision--language agents.
ReGRPO builds a Structured Reflective Data Engine and applies a GRPO-based reflection protocol to optimize reflection and action generation jointly in tool-using trajectories.
Our method keeps the standard GRPO optimizer, with the main design changes in structured reflection representation and the training/deployment protocol.

On the GTA and GAIA benchmarks, our approach consistently outperforms action-level and SFT-only baselines under the same backbone/tool setting.
The ablation results show complementary gains: RoT SFT provides a strong foundation, structured RL contributes further improvements, and optional verifier distillation/reward adds further but smaller gains.
Overall, ReGRPO improves answer accuracy and tool correctness over MAT-AGENT and SPORT while using the same VLM backbone and tool suite.

\subsubsection*{Acknowledgements.}
This project is supported by the Ministry of Education, Singapore, under its Academic Research Fund Tier 2 (Award No: MOE-T2EP20124-0012).

\clearpage

\begingroup
\sloppy
\bibliographystyle{splncs04}
\bibliography{main}
\endgroup

\appendix
\setcounter{section}{0}
\renewcommand{\thesection}{A\arabic{section}}
\renewcommand{\theHsection}{appendix.\arabic{section}}

\definecolor{codebg}{HTML}{F4F4F6}
\definecolor{cardhead}{HTML}{2E5E9E}
\newcommand{\promptcard}[2]{%
  \par\smallskip
  \begingroup\setlength{\fboxsep}{5pt}\setlength{\parskip}{0pt}%
  \noindent\colorbox{cardhead}{\parbox{\dimexpr\linewidth-2\fboxsep\relax}{\footnotesize\textcolor{white}{\textbf{#1}}}}\par\nointerlineskip
  \noindent\colorbox{codebg}{\parbox{\dimexpr\linewidth-2\fboxsep\relax}{#2}}\par
  \endgroup
  \smallskip}

\clearpage
\newpage
\setcounter{page}{1}
\setcounter{linenumber}{1}

\begin{center}
  {\LARGE \textbf{Appendix}}
\end{center}
\vspace{1em}

\section{Algorithmic Details}
\label{sec:algorithms}

In this section we provide pseudo code and implementation details for Reflection-Augmented Group Relative Policy Optimization (ReGRPO).

\subsection{ReGRPO Training}

We assume a dataset of task states. For each state $s_i$, we sample a group of trajectories and optimize a verifier-aware reward during training. The verifier term is used only in training; inference is strictly zero-verifier.

\begin{algorithm}[h]
  \caption{ReGRPO Training Loop (Verifier-Aware Reward)}
  \label{alg:r_grpo}
  \begin{algorithmic}[1]
    \REQUIRE Dataset $\mathcal{D}$, policy $\pi_\theta$, ref policy $\pi_{\text{ref}}$, coefficients $\lambda_{\text{exec}},\lambda_{\text{val}},\eta$, KL weight $\beta$, verifier weights $w_a,w_g,w_p$, group size $K$.
    \WHILE{not converged}
    \STATE Sample batch of states $S \subset \mathcal{D}$
    \FORALL{state $s \in S$}
    \STATE Construct a group of $K$ candidate trajectories $\{\tau^k\}_{k=1}^K$
    \FOR{$k=1$ \TO $K$}
    \STATE Replay $\tau^k$ to collect success signal and tool observations
    \STATE Compute verifier subscores $(s_a^k,s_g^k,s_p^k)$
    \STATE $V(\tau^k) = w_a s_a^k + w_g s_g^k + w_p s_p^k$
    \STATE Compute reward $R(\tau^k) = \lambda_{\text{exec}}\cdot \text{Success}(\tau^k) + \lambda_{\text{val}}\cdot V(\tau^k) - \eta \cdot C(\tau^k)$
    \ENDFOR
    \STATE Compute baseline $\bar{R} = \frac{1}{K}\sum_{k=1}^K R(\tau^k)$
    \STATE Compute advantages $A^k = R(\tau^k) - \bar{R}$
    \STATE Accumulate gradients $\nabla_\theta \mathcal{L} = -\sum_k A^k \nabla \log \pi_\theta(\tau^k) + \beta\,\mathrm{KL}(\pi_\theta\|\pi_{\text{ref}})$
    \ENDFOR
    \STATE Update $\theta$
    \ENDWHILE
  \end{algorithmic}
\end{algorithm}

\subsection{Reward Coefficients and Group Size}

Table~\ref{tab:reward_coeffs} lists the default coefficients used in our reported setting.

\begin{table}[h]
  \centering
  \caption{Default ReGRPO coefficients. In deployment we set $\lambda_{\text{val}}=0$ and do not call external verifiers.}
  \label{tab:reward_coeffs}
  \scriptsize
  \begin{tabular}{lcc}
    \hline
    \textbf{Symbol} & \textbf{Meaning} & \textbf{Value} \\
    \hline
    $\lambda_{\text{exec}}$ & execution success reward weight & 1.0 \\
    $\lambda_{\text{val}}$ & verifier reward weight (training only) & 0.3 \\
    $\eta$ & reflection cost penalty & 0.1 \\
    $\beta$ & KL penalty weight & 0.3 \\
    $w_a$ & argument correctness weight & 0.25 \\
    $w_g$ & grounding consistency weight & 0.50 \\
    $w_p$ & task progress weight & 0.25 \\
    $K$ & candidate trajectories per group & 5 \\
    \hline
  \end{tabular}
\end{table}

\subsection{Inference Trigger Implementation}
\label{sec:hard_failure_sentinel}

We use a deterministic and lightweight trigger that avoids per-tool feature engineering.
For each step, tool output is normalized as
\begin{equation}
  \hat{o}_i = \{\texttt{status},\texttt{payload},\texttt{meta}\},
\end{equation}
where \texttt{status} records runtime errors and \texttt{payload} stores normalized content.
The trigger is
\begin{equation}
  g_i = \mathbf{1}\{\texttt{ToolError}(\hat{o}_i) \lor \texttt{EmptyObs}(\hat{o}_i) \lor u_i < \kappa_i\}.
\end{equation}
Hard failures are captured by \texttt{ToolError}/\texttt{EmptyObs}; silent failures are captured by low policy confidence
\begin{equation}
  u_i = \exp\!\left(\frac{1}{|a_i^{(0)}|}\sum_j \log \pi_\theta(a_{i,j}^{(0)}\mid x_i,h_i)\right).
\end{equation}
To avoid offline calibration and per-tool thresholds, we use an online adaptive threshold
\begin{equation}
  \kappa_i = \frac{1}{\max(1,i-1)}\sum_{j=1}^{i-1} u_j.
\end{equation}
For $i=1$, confidence-based triggering is disabled and only hard-failure checks are used.

\begin{algorithm}[t]
  \caption{Minimal Deterministic Trigger (Zero-Verifier)}
  \label{alg:det_gate}
  \begin{algorithmic}[1]
    \REQUIRE Context $x_i$, tentative action $a_i^{(0)}$, tool output $o_i^{(0)}$, previous confidences $\{u_j\}_{j<i}$
    \STATE Normalize $o_i^{(0)}$ to canonical schema $\hat{o}_i$
    \STATE Compute confidence $u_i$ from action log-probabilities
    \STATE Compute adaptive threshold $\kappa_i$ as mean confidence of previous steps
    \STATE $g_i \leftarrow \mathbf{1}\{\texttt{ToolError}(\hat{o}_i) \lor \texttt{EmptyObs}(\hat{o}_i) \lor (i>1 \land u_i < \kappa_i)\}$
    \STATE If $g_i=1$ and local-reflection-count $<1$, run one block $a_i^{(0)}\!\rightarrow\!o_i^{(0)}\!\rightarrow\!z_i\!\rightarrow\!a_i^{(1)}\!\rightarrow\!o_i^{(1)}$
    \ENSURE Trigger decision $g_i$
  \end{algorithmic}
\end{algorithm}

This design keeps deployment deterministic, simple, and reproducible while reducing manual tuning.

\section{Structured Reflective Data Engine Statistics}
\label{sec:rot_stats}

{\sloppy Table~\ref{tab:rot_perturb_dist} reports corpus statistics over the synthesized Reflection-on-Thought (RoT) data ($16{,}552$ records). Each record is labeled with one of the four \texttt{ErrorType} categories used throughout the paper (Sec.~\ref{subsec:mmcot}): \texttt{ArgInvalid}, \texttt{ToolMismatch}, \texttt{InfoInsufficient}, and \texttt{GroundingDrift}. This four-way taxonomy is the reflection label set: \texttt{ArgInvalid} dominates ($45.4\%$), reflecting that corrupted-argument failures (wrong path/index/span) are the most frequent recoverable error in the source trajectories, followed by tool-capability mismatches ($24.7\%$), insufficient-information queries ($19.7\%$), and grounding drift ($10.1\%$). The corpus is multi-modal: $37.0\%$ of records carry an image, while the remaining $63.0\%$ are text-only.\par}

\begin{table}[!h]
  \centering
  \caption{\texttt{ErrorType} distribution over the synthesized Reflection-on-Thought corpus ($16{,}552$ records). The reflection label set is the four-way taxonomy used throughout the paper. Across the corpus, $37.0\%$ of records carry an image and $63.0\%$ are text-only.}
  \label{tab:rot_perturb_dist}
  \scriptsize
  \begin{tabular}{lcc}
    \hline
    \textbf{ErrorType} & \textbf{Count} & \textbf{Share (\%)} \\
    \hline
    \texttt{ArgInvalid} & 7{,}518 & 45.4 \\
    \texttt{ToolMismatch} & 4{,}096 & 24.7 \\
    \texttt{InfoInsufficient} & 3{,}263 & 19.7 \\
    \texttt{GroundingDrift} & 1{,}675 & 10.1 \\
    \hline
    \textbf{Total} & \textbf{16{,}552} & \textbf{100.0} \\
    \hline
  \end{tabular}
\end{table}

\begin{table}[!h]
  \centering
  \caption{Tool used in the perturbed action over the synthesized Reflection-on-Thought corpus (top entries). The distribution is dominated by the image-QA, search, and file-inspection tools that appear most often in the source trajectories.}
  \label{tab:rot_tool_dist}
  \scriptsize
  \begin{tabular}{lc}
    \hline
    \textbf{Tool} & \textbf{Share (\%)} \\
    \hline
    \texttt{ask\_search\_agent} & 32.1 \\
    \texttt{visualizer} & 22.5 \\
    \texttt{inspect\_file\_as\_text} & 19.4 \\
    \texttt{search} & 10.9 \\
    \texttt{image\_generator} & 5.9 \\
    \texttt{objectlocation} & 3.8 \\
    \texttt{facedetection} & 2.8 \\
    \texttt{image\_edit} & 1.9 \\
    \texttt{segmentation} & 0.7 \\
    \hline
  \end{tabular}
\end{table}

\begin{table}[!h]
  \centering
  \caption{Component contribution of the Structured Reflective Data Engine and reflection-aware RL.}
  \label{tab:rot_component_contrib}
  \scriptsize
  \begin{tabular}{lcc}
    \hline
    \textbf{Variant} & \textbf{GTA AnsAcc} & \textbf{GAIA AnsAcc} \\
    \hline
    MM-Traj SFT (MAT-AGENT) & 53.85 & 16.97 \\
    + RoT SFT & 58.59 & 19.03 \\
    + ReGRPO core (default) & \textbf{67.66} & \textbf{23.35} \\
    \hline
  \end{tabular}
\end{table}

\section{Verifier Subscores and Offline Teacher}
\label{sec:verifier_prompt}

We stress that no LLM verifier is queried inside the RL loop. The verifier value
$V$ used by the $\lambda_{\text{val}} V$ reward term is computed
\emph{deterministically} from each candidate and the active record's RoT metadata,
so $V$ is a function of the candidate, not of any network or external model call:
$$ V = w_a s_a + w_g s_g + w_p s_p, \quad w_g \ge w_a, w_p, \quad s_a,s_g,s_p \in [0,1], $$
with default weights $(w_a, w_g, w_p) = (0.25, 0.50, 0.25)$. The three subscores are
derived as follows (see Fig.~\ref{fig:verifier_prompt}):
\begin{itemize}
\item \textbf{$s_p$ (plan validity):} $s_p = 1$ iff the candidate's normalized primary
  tool and first argument match the stored \texttt{corrected\_action} signature, else $0$.
\item \textbf{$s_a$ (answer consistency):} $s_a = 1$ iff the candidate's replay succeeds
  (its terminal action agrees with the group's correct answer), else $0$.
\item \textbf{$s_g$ (grounding):} $s_g = 1$ iff the candidate carries a reflection whose
  \texttt{evidence} is text-grounded in the stored \texttt{failed\_observation}
  \emph{and} $s_p > 0$, else $0$. There is intentionally no fallback to $s_p$ for
  reflection-less candidates, so $V$ rewards grounded reflection above bare plan repair.
\end{itemize}
A GPT-4o teacher is used \emph{only offline}, to synthesize the RoT reflections in the
data engine (Sec.~\ref{sec:prompts}); it is never called during RL training or at inference.

\begin{figure}[!h]
  \centering
  \scriptsize
  \promptcard{Deterministic Verifier Subscores~\textnormal{(no LLM in the loop)}}{%
    \footnotesize\textbf{Inputs} \textnormal{(from the active RoT record):}\par\smallskip
    {\ttfamily\scriptsize\linespread{1.3}\selectfont
    candidate.code \quad// the sampled action under evaluation\\
    meta.corrected\_action \quad// gold action signature $a^{*}$\\
    meta.failed\_observation,\ candidate.reflection.evidence\\
    candidate.success \quad// replay success flag\par}
    \smallskip\footnotesize\textbf{Computation:}\par\smallskip
    {\ttfamily\scriptsize\linespread{1.3}\selectfont
    s\_p = 1 if signature(candidate.code) == signature(corrected\_action) else 0\\
    s\_a = 1 if candidate.success else 0\\
    grounded = evidence\_grounded(reflection.evidence, failed\_observation)\\
    s\_g = 1 if (grounded and s\_p > 0) else 0\\
    V = 0.25*s\_a + 0.50*s\_g + 0.25*s\_p\par}
    \smallskip\footnotesize\textbf{No external model query; deterministic given the metadata.}}
\caption{Teacher-derived verifier subscores. At training time $(s_a, s_g, s_p)$ are computed deterministically from the teacher's RoT metadata via tool/argument-signature matching, the replay success flag, and a grounded-reflection check ($s_g$ requires grounded evidence and $s_p>0$); no GPT-4o (or any LLM) is queried in the RL loop. The GPT-4o teacher is used only offline to synthesize the RoT reflections.}
\label{fig:verifier_prompt}
\end{figure}

\section{Tool Suite}
\label{sec:tools}

We use the standard MAT-AGENT tool suite:
\begin{itemize}
\item \textbf{Web search}: \texttt{searchinformation}, \texttt{visit}, \texttt{webqa}.
\item \textbf{Image QA}: \texttt{image\_qa}.
\item \textbf{File inspector}: \texttt{pdf\_span}, \texttt{table\_reader}.
\item \textbf{Object localization}: \texttt{object\_loc}.
\item \textbf{Python sandbox}: \texttt{python\_exec}.
\end{itemize}

\section{Prompt Templates}
\label{sec:prompts}

\subsection{Reflection Generation Prompt}

The Reflection-on-Thought (RoT) corpus is synthesized offline by a teacher LLM (GPT-4o).
For each source step, the teacher is asked to produce, under a single requested
\texttt{ErrorType}, one realistic near-miss \emph{failed action}, one faithful \emph{failed observation}
in the target tool's native output format, and one structured \emph{reflection} triple.
The system prompt fixes the contract and enforces strict JSON output:

\promptcard{System Prompt}{%
\small
You are a strict data generation engine for ReGRPO.
Produce one realistic near-miss failed action, one faithful failed observation,
and one Reflection-of-Thought triple.
Return STRICT JSON only. Do not include markdown, prose, comments, or extra keys.
The failed action must keep the same Thought plus Code shape as the provided correct action,
but it must be broken according to the requested \texttt{ErrorType}.
The failed observation must look like the named tool's real output format and must be
consistent with the failed action.
The reflection evidence must quote a concrete token or phrase from \texttt{failed\_observation}.}

{\sloppy The user message is a JSON payload that supplies the task, the source step, the
\texttt{requested\_error\_type} drawn from the four-way taxonomy
$\{$\texttt{ArgInvalid}, \texttt{ToolMismatch}, \texttt{InfoInsufficient}, \texttt{GroundingDrift}$\}$,
a per-error-type guidance string, a per-tool failure-observation template, the history before
the step, and the (teacher-only) correct action. It also fixes the exact output schema the
teacher must emit:\par}

\promptcard{User Message~\textnormal{(abridged JSON)}}{%
\ttfamily\scriptsize\linespread{1.3}\selectfont
\{\\
\ \ "task": ..., "source\_id": ..., "step\_index": ...,\\
\ \ "target\_tool": ..., "requested\_error\_type": <one of ArgInvalid|ToolMismatch|\\
\ \ InfoInsufficient|GroundingDrift>, "error\_type\_guidance": [...],\\
\ \ "failure\_observation\_template": [...], "history\_before\_step": [...],\\
\ \ "correct\_action\_teacher\_only": [...],\\
\ \ "output\_schema": \{\\
\ \ \ \ "failed\_action": "<Thought+Code, same shape as correct but broken>",\\
\ \ \ \ "failed\_observation": "<realistic failure in the tool's output format>",\\
\ \ \ \ "reflection": \{\\
\ \ \ \ \ \ "error\_type": <the requested ErrorType>,\\
\ \ \ \ \ \ "evidence": "<quote a token or phrase from failed\_observation>",\\
\ \ \ \ \ \ "fix\_plan": "<concrete corrective strategy>"\\
\ \ \ \ \}\\
\ \ \}\\
\}}

{\sloppy The per-error-type guidance instantiates each label concretely, e.g.\ \texttt{GroundingDrift}
``shift the referenced region, object, crop, or visual target while preserving the tool shape'';
\texttt{ToolMismatch} ``replace the tool with a different, unsuitable tool for the same intent'';
\texttt{ArgInvalid} ``corrupt an argument such as a path, page, bbox, object name, query, size, or format'';
and \texttt{InfoInsufficient} ``drop or underspecify necessary context so the tool cannot retrieve
enough information.''
Every emitted record passes a deterministic validator that, among other checks, requires
\texttt{reflection.evidence} to be text-grounded in \texttt{failed\_observation} (a literal token
overlap heuristic) and rejects any record whose \texttt{error\_type} lies outside the taxonomy.\par}

\section{Base VQA Ability Is Preserved}
\label{sec:base_vqa}

A natural concern is whether reflection-augmented RL erodes the controller's underlying vision--language ability. Evaluating the base model and the \ours{}-trained model on MMBench, we find that \ours{} preserves base VQA ability ($85.0 \to 84.25$, a $0.75$-point change), so learning structured reflection does not come at the cost of general VQA competence.

\section{Qualitative Study and Case Studies}
\label{sec:qualitative}

We present trajectory-level case studies under the same inference setup as Sec.~\ref{subsec:inference}: inference is single-path and zero-verifier, and at each step we apply the deterministic gate
\(g_i=\mathbf{1}\{\texttt{ToolError}(\hat{o}_i) \lor \texttt{EmptyObs}(\hat{o}_i) \lor (i>1 \land u_i<\kappa_i)\}\),
allowing at most one local reflection-correction block per step.
All trajectories and records are copied \emph{verbatim} from the \ours{} evaluation caches and the Structured Reflective Data Engine corpus; no trajectory is paraphrased or synthesized, and long fields are truncated with ``[\dots]'' for readability only.

\subsection{Success Trajectories}
\label{sec:success_traj}

We first show two correct end-to-end trajectories from the GTA evaluation cache. Correctness is checked with GTA's official substring rule (for every gold list, at least one gold string is a case-insensitive substring of the final answer). Trajectories use the runtime MAT-AGENT tool names (\texttt{visualizer}, \texttt{ask\_search\_agent}), which correspond respectively to the image-QA and web-search tools.

\begin{figure}[h]
\centering
\includegraphics[width=0.45\linewidth]{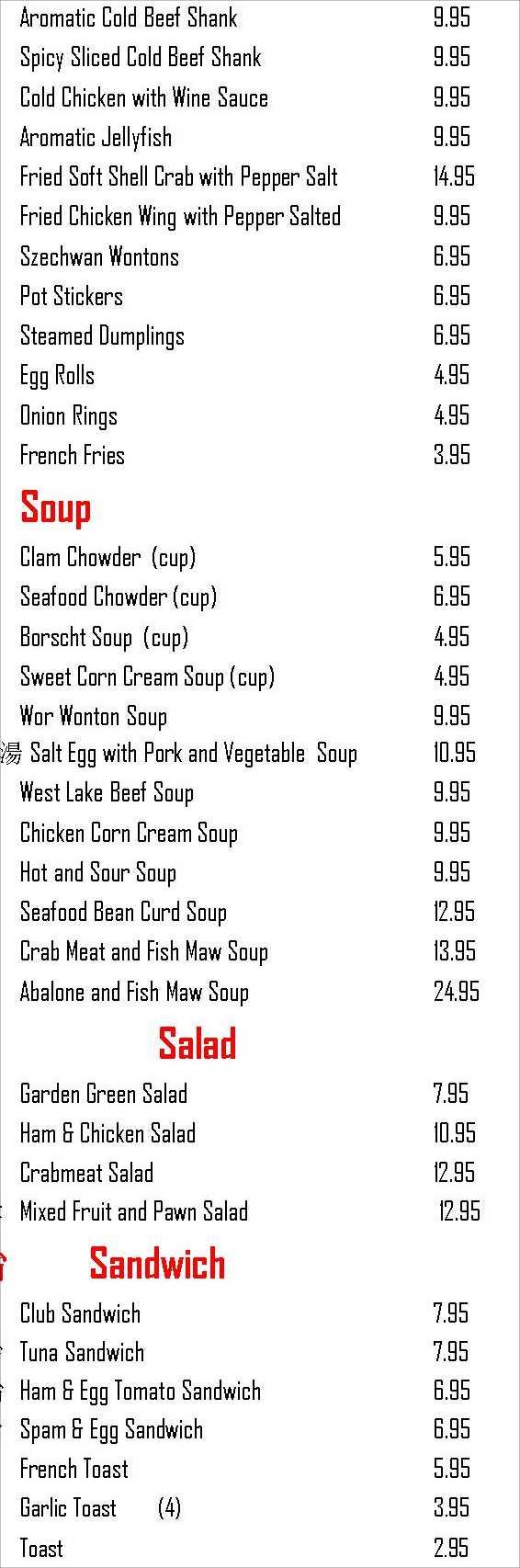}
\caption{Input image for Case A1 (\texttt{image\_318.jpg}): the menu the agent grounds and reads before solving the constrained selection.}
\label{fig:case_a1_img}
\end{figure}

\paragraph{Case A1: GTA multi-step tool use --- constrained menu selection.}
\textbf{Task:} ``Choose only one dish in Salad and one dish in Sandwich, which combination can be no more than 10.9 yuan in total?''\\
Input: \texttt{image\_318.jpg} (a menu, Figure~\ref{fig:case_a1_img}). Gold: \{Toast, Garden Green Salad\}.

\begin{table*}[t]
  \centering
  \caption{Case A1 (GTA, \texttt{\_\_pf2}, id~137). Image grounding (reading the structured menu) followed by exact constrained optimization in code: the only pair summing to $\le 10.9$ is found correctly. Verified by the GTA substring rule.}
  \label{tab:case_a1}
  \scriptsize
  \begin{tabular}{p{0.06\linewidth}p{0.30\linewidth}p{0.26\linewidth}p{0.30\linewidth}}
    \hline
    \textbf{Step} & \textbf{Action / Reflection} & \textbf{Observation} & \textbf{Gate \& Outcome} \\
    \hline
    1 & \texttt{visualizer} --- \emph{question:} ``Extract the Salad and Sandwich sections from this menu.''; \emph{image\_path:} \texttt{.cache/image\_318.jpg} & Parsed menu. Salad: Garden Green 7.95, Ham \& Chicken 10.95, [\dots]; Sandwich: [\dots] French Toast 5.95, Garlic Toast 3.95, Toast 2.95 & No hard failure ($g_1{=}0$); the menu is grounded from the image. \\
    2 & \texttt{python\_exec} --- enumerate Salad $\times$ Sandwich pairs with total $\le 10.9$ yuan & Valid combinations: (Garden Green Salad, Toast, 10.9) & $g_2{=}0$; exact constrained search over the grounded prices. \\
    3 & \texttt{final\_answer} --- (Garden Green Salad, Toast, 10.9) & --- & Both gold strings appear; \checkmark\ correct ($7.95{+}2.95{=}10.90$). \\
    \hline
  \end{tabular}
\end{table*}

\begin{figure}[h]
\centering
\includegraphics[width=0.45\linewidth]{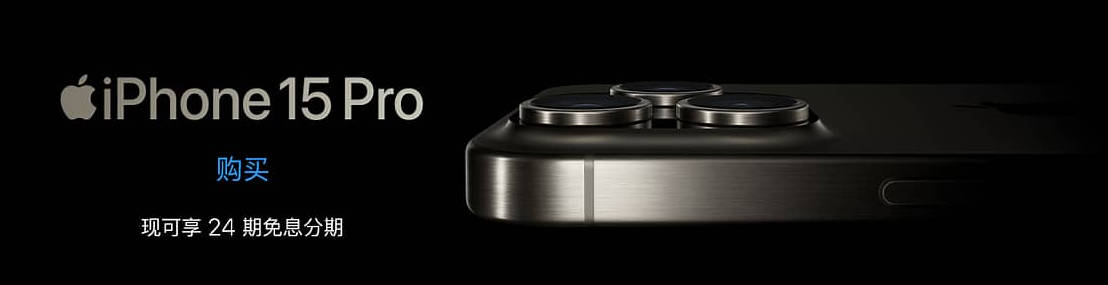}
\caption{Input image for Case A2 (\texttt{image\_417.jpg}): the depicted product the agent recognizes and resolves to its manufacturer before retrieving the CEO.}
\label{fig:case_a2_img}
\end{figure}

\paragraph{Case A2: GTA search $+$ reason --- image $\rightarrow$ entity $\rightarrow$ web fact.}
\textbf{Task:} ``Who is the CEO of this company?''\\
Input: \texttt{image\_417.jpg} (an iPhone~15~Pro, Figure~\ref{fig:case_a2_img}). Gold: \{Tim Cook\}.

\begin{table*}[t]
  \centering
  \caption{Case A2 (GTA, \texttt{\_\_pf2}, id~213). A search-and-reason chain: recognize the depicted product, resolve it to the manufacturer, then retrieve and report the CEO. The agent selects the search tool rather than an image tool.}
  \label{tab:case_a2}
  \scriptsize
  \begin{tabular}{p{0.06\linewidth}p{0.30\linewidth}p{0.26\linewidth}p{0.30\linewidth}}
    \hline
    \textbf{Step} & \textbf{Action / Reflection} & \textbf{Observation} & \textbf{Gate \& Outcome} \\
    \hline
    1 & \texttt{ask\_search\_agent} --- \emph{query:} ``Who is the CEO of the company that produces the iPhone 15 Pro?'' & ``The iPhone 15 Pro is produced by Apple Inc. The CEO of Apple is Tim Cook.'' & $g_1{=}0$; product $\rightarrow$ manufacturer (Apple) $\rightarrow$ CEO. \\
    2 & \texttt{final\_answer} --- ``The CEO of Apple, which produces the iPhone 15 Pro, is Tim Cook.'' & --- & Final answer contains gold \texttt{Tim Cook}; \checkmark\ correct. \\
    \hline
  \end{tabular}
\end{table*}

\clearpage
\subsection{Failure Modes and Reflective Recovery}
\label{sec:synth_rot}

The four records below are copied \emph{verbatim} from the Structured Reflective Data Engine corpus (Sec.~\ref{subsec:mmcot}) and cover the four \texttt{ErrorType} categories used throughout the paper: \texttt{ArgInvalid}, \texttt{ToolMismatch}, \texttt{InfoInsufficient}, and \texttt{GroundingDrift}. For one error type each, the record shows the task, the failed (perturbed) action $a_i^{fail}$, the resulting failure observation $o_i^{fail}$, the synthesized structured Reflection-on-Thought $z_i$ (\texttt{ErrorType} / \texttt{Evidence} / \texttt{FixPlan}, the schema of Sec.~\ref{subsec:mmcot}), and the corrected action $a_i^{*}$. During supervised fine-tuning the reflection is emitted in-context as a \texttt{Reflection:} block immediately before the corrected action, so the model maximizes $P(z_i, a_i^{*}\mid x_i, a_i^{fail}, o_i^{fail})$.

\begin{table*}[t]
  \centering
  \caption{Case B1 --- \texttt{ArgInvalid} (record \texttt{00WXZu4s\_rot\_s2\_ArgInvalid}). The reflection notices the observation describes traffic signs (not the target image), traces the cause to a corrupted path (\texttt{...\_jpg} with an extra underscore instead of \texttt{....jpg}), and repairs the argument to the known-good path.}
  \label{tab:rot_b1}
  \scriptsize
  \begin{tabular}{p{0.16\linewidth}p{0.76\linewidth}}
    \hline
    \textbf{Task} & ``I would like to know the most common colors in the image. Please summarize the findings.'' \quad(image \texttt{llava/llava\_pretrain/images/00022/000229128.jpg}) \\
    \hline
    \textbf{Failed action} & \texttt{visualizer} with \emph{question} ``What are the most common colors in this image?'' and \emph{image\_path} \texttt{data/tongagent/llava/llava\_pretrain/images/00022/000229128\_jpg} (note: ends in \texttt{\_jpg} instead of \texttt{.jpg}). \\
    \textbf{Failed obs.} & ``The image mainly contains bright red and yellow traffic signs against a gray urban street background.'' \\
    \hline
    \multicolumn{2}{l}{\textbf{Reflection (Reflection-on-Thought $z_i$, emitted before the correction)}} \\
    \quad\textbf{ErrorType} & \texttt{ArgInvalid} \\
    \quad\textbf{Evidence} & ``bright red and yellow traffic signs against a gray urban street background'' \\
    \quad\textbf{FixPlan} & The \texttt{visualizer} call used an invalid \texttt{image\_path} with the wrong file extension and an extra underscore. Reuse the known-good path \texttt{data/tongagent/llava/llava\_pretrain/images/00022/000229128.jpg} exactly, ensuring the extension and spelling match the original image source before asking about the most common colors. \\
    \hline
    \textbf{Corrected action} & \texttt{visualizer} with the same \emph{question} and \emph{image\_path}\,$=$\,\texttt{image\_path} (the known-good path variable). \\
    \hline
  \end{tabular}
\end{table*}

\begin{table*}[t]
  \centering
  \caption{Case B2 --- \texttt{ToolMismatch} (record \texttt{00\_tFZUB\_rot\_s0\_ToolMismatch}). The reflection reads the tool's own error (``does not access historical temperature records''), recognizes a capability mismatch, and switches from \texttt{image\_generator} to \texttt{ask\_search\_agent} --- a fact-retrieval goal routed to a search tool.}
  \label{tab:rot_b2}
  \scriptsize
  \begin{tabular}{p{0.16\linewidth}p{0.76\linewidth}}
    \hline
    \textbf{Task} & ``I need to find the average temperature of the coldest week in January 2023 in New York City. Could you look that up for me and also provide the dates of that week?'' \\
    \hline
    \textbf{Failed action} & \texttt{image\_generator} with \emph{prompt} ``Infographic showing the average temperature of the coldest week in January 2023 in New York City, labeled with the dates of that week''. \\
    \textbf{Failed obs.} & ``\texttt{image\_generator} output: unable to provide factual numeric weather data or specific dates; this tool only generates images based on text prompts and does not access historical temperature records.'' \\
    \hline
    \multicolumn{2}{l}{\textbf{Reflection (Reflection-on-Thought $z_i$, emitted before the correction)}} \\
    \quad\textbf{ErrorType} & \texttt{ToolMismatch} \\
    \quad\textbf{Evidence} & ``this tool only generates images based on text prompts and does not access historical temperature records'' \\
    \quad\textbf{FixPlan} & Use a search-capable tool (\eg, \texttt{search} or \texttt{ask\_search\_agent}) with a precise query about January 2023 NYC temperatures to retrieve factual historical weather data, then compute or extract the average temperature and corresponding dates of the coldest week. \\
    \hline
    \textbf{Corrected action} & \texttt{ask\_search\_agent} with \emph{query} ``What was the average temperature of the coldest week in January 2023 in New York City and what were the dates of that week?'' \\
    \hline
  \end{tabular}
\end{table*}

\clearpage
\begin{table*}[t]
  \centering
  \caption{Case B3 --- \texttt{InfoInsufficient} (record \texttt{09ysaaTv\_rot\_s0\_InfoInsufficient}). A generic ``what is shown'' prompt returned only a scene description (no artist/year); the reflection rewrites the query to ask the tool for exactly the requested facts while keeping the correct image path.}
  \label{tab:rot_b3}
  \scriptsize
  \begin{tabular}{p{0.16\linewidth}p{0.76\linewidth}}
    \hline
    \textbf{Task} & ``I've attached a photo of a painting. Can you tell me the artist's name and the year it was created?'' \quad(image \texttt{wikiart/images/fyodor-vasilyev\_village-1.jpg}) \\
    \hline
    \textbf{Failed action} & \texttt{visualizer} with \emph{question} ``What is shown in this painting?'' and \emph{image\_path} \texttt{data/tongagent/wikiart/images/fyodor-vasilyev\_village-1.jpg}. \\
    \textbf{Failed obs.} & ``The image shows a rural landscape with small houses, trees, and a cloudy sky.'' \\
    \hline
    \multicolumn{2}{l}{\textbf{Reflection (Reflection-on-Thought $z_i$, emitted before the correction)}} \\
    \quad\textbf{ErrorType} & \texttt{InfoInsufficient} \\
    \quad\textbf{Evidence} & ``The image shows a rural landscape with small houses, trees, and a cloudy sky.'' \\
    \quad\textbf{FixPlan} & Explicitly ask the visualizer for the artist's name and the year the painting was created, rather than a generic content description, while keeping the same image path. \\
    \hline
    \textbf{Corrected action} & \texttt{visualizer} with \emph{question} ``What is the name of the artist and the year this painting was created?'' and the same \emph{image\_path}. \\
    \hline
  \end{tabular}
\end{table*}

\begin{table*}[t]
  \centering
  \caption{Case B4 --- \texttt{GroundingDrift} (record \texttt{0FLZe2lb\_rot\_s0\_GroundingDrift}). The reflection detects that the query drifted to the cutting board and utensils (``the smoothie glass is not described'') and re-anchors the visual question to the smoothie itself, so the extracted ingredients feed the actual calorie estimate.}
  \label{tab:rot_b4}
  \scriptsize
  \begin{tabular}{p{0.16\linewidth}p{0.76\linewidth}}
    \hline
    \textbf{Task} & ``How many calories are in the smoothie shown in this image?'' \quad(image \texttt{coco/train2017/000000043093.jpg}) \\
    \hline
    \textbf{Failed action} & \texttt{visualizer} with \emph{question} ``What fruits and utensils are visible on the table in this image?'' and \emph{image\_path} \texttt{data/tongagent/coco/train2017/000000043093.jpg}. \\
    \textbf{Failed obs.} & ``There is a wooden cutting board with a sliced banana, a knife, and some scattered granola on the table; a glass jar of yogurt is also visible next to the board, but the smoothie glass is not described.'' \\
    \hline
    \multicolumn{2}{l}{\textbf{Reflection (Reflection-on-Thought $z_i$, emitted before the correction)}} \\
    \quad\textbf{ErrorType} & \texttt{GroundingDrift} \\
    \quad\textbf{Evidence} & ``a wooden cutting board with a sliced banana, a knife'' \\
    \quad\textbf{FixPlan} & Refocus the question and visual attention from the general table setup and cutting board to the actual smoothie glass. Ask the visualizer specifically about the contents of the smoothie cup or glass (color, texture, visible toppings) rather than surrounding fruits and utensils, so the extracted ingredients correspond directly to the drink whose calories we need to estimate. \\
    \hline
    \textbf{Corrected action} & \texttt{visualizer} with \emph{question} ``What ingredients are visible in this smoothie image?'' and the same \emph{image\_path}. \\
    \hline
  \end{tabular}
\end{table*}

\end{document}